\documentclass{article}

\usepackage{microtype}
\usepackage{graphicx}
\usepackage{subfigure}
\usepackage{booktabs} 

\usepackage{overpic}

\usepackage{hyperref}

\usepackage{bera}
\usepackage{listings}
\usepackage{xcolor}
\usepackage{textcomp}
\usepackage{color}

\colorlet{punct}{red!60!black}
\definecolor{background}{HTML}{EEEEEE}
\definecolor{delim}{RGB}{20,105,176}
\colorlet{numb}{magenta!60!black}

\lstdefinelanguage{json}{
    basicstyle=\small\ttfamily,
    numbers=left,
    numberstyle=\scriptsize,
    stepnumber=1,
    numbersep=8pt,
    showstringspaces=false,
    breaklines=true,
    frame=lines,
    backgroundcolor=\color{background},
    literate=
     *{0}{{{\color{numb}0}}}{1}
      {1}{{{\color{numb}1}}}{1}
      {2}{{{\color{numb}2}}}{1}
      {3}{{{\color{numb}3}}}{1}
      {4}{{{\color{numb}4}}}{1}
      {5}{{{\color{numb}5}}}{1}
      {6}{{{\color{numb}6}}}{1}
      {7}{{{\color{numb}7}}}{1}
      {8}{{{\color{numb}8}}}{1}
      {9}{{{\color{numb}9}}}{1}
      {:}{{{\color{punct}{:}}}}{1}
      {,}{{{\color{punct}{,}}}}{1}
      {\{}{{{\color{delim}{\{}}}}{1}
      {\}}{{{\color{delim}{\}}}}}{1}
      {[}{{{\color{delim}{[}}}}{1}
      {]}{{{\color{delim}{]}}}}{1},
}

\usepackage{amsmath,amsfonts,bm}

\def\eqref#1{equation~\ref{#1}}

\def\1{\bm{1}}

\DeclareMathAlphabet{\mathsfit}{\encodingdefault}{\sfdefault}{m}{sl}
\SetMathAlphabet{\mathsfit}{bold}{\encodingdefault}{\sfdefault}{bx}{n}

\usepackage{hyperref}
\usepackage{url}
\usepackage{xcolor}
\usepackage{graphicx}
\usepackage[english]{babel}
\usepackage{soul}

\usepackage[toc,page,header]{appendix}
\usepackage{titletoc}

\definecolor{darkgreen}{rgb}{0,0.6,0}
\definecolor{darkred}{rgb}{0.7,0.0,0}
\definecolor{darkblue}{rgb}{0,0.0,0.6}
\definecolor{magenta}{rgb}{0.8,0.1,0.8}
\definecolor{darksomething}{rgb}{0,0.4,0.6}

\newcommand{\ncom}[1]{}

\definecolor{darkgreen}{rgb}{0,0.6,0}
\definecolor{darkred}{rgb}{0.7,0.0,0}
\definecolor{darkblue}{rgb}{0,0.0,0.6}
\definecolor{magenta}{rgb}{0.8,0.1,0.8}
\definecolor{darksomething}{rgb}{0,0.4,0.6}

\PassOptionsToPackage{numbers, compress}{natbib}
\usepackage[preprint]{neurips_2020}

\usepackage[utf8]{inputenc} 
\usepackage[T1]{fontenc}    
\usepackage{hyperref}       
\usepackage{url}            
\usepackage{booktabs}       
\usepackage{amsfonts}       
\usepackage{nicefrac}       
\usepackage{microtype}      

\title{Training Learned Optimizers with Randomly Initialized Learned Optimizers}

\author{%
  Luke Metz, C. Daniel Freeman, Niru Maheswaranathan, Jascha Sohl-Dickstein\\
  Google Research, Brain Team\\
  \texttt{\{lmetz, cdfreeman, nirum, jaschasd\}@google.com}
}

\begin{document}

\maketitle
\vspace{-20pt}
\section{Abstract}
Learned optimizers are increasingly effective, with performance exceeding that of hand designed optimizers such as Adam~\citep{kingma2014adam}
on specific tasks \citep{metz2019understanding}. 
Despite the potential gains available,
in current work the meta-training (or `outer-training') of the learned optimizer is performed by a hand-designed optimizer, or by an optimizer trained by a hand-designed optimizer \citep{metz2020tasks}. 
We show that a population of randomly initialized learned optimizers can be used to train themselves from scratch in an online fashion, without resorting to a hand designed optimizer in any part of the process.
A form of population based training \citep{jaderberg2017population} is used to orchestrate this self-training. 
Although the randomly initialized optimizers initially make slow progress, as they improve they experience a positive feedback loop, and become rapidly more effective at training themselves. 
We believe feedback loops of this type, where an optimizer improves itself, will be important and powerful in the future of machine learning. These methods not only provide a path towards increased performance, but more importantly relieve research and engineering effort.

\begin{figure}[bh]
    \centering
    \includegraphics[width=0.9\textwidth]{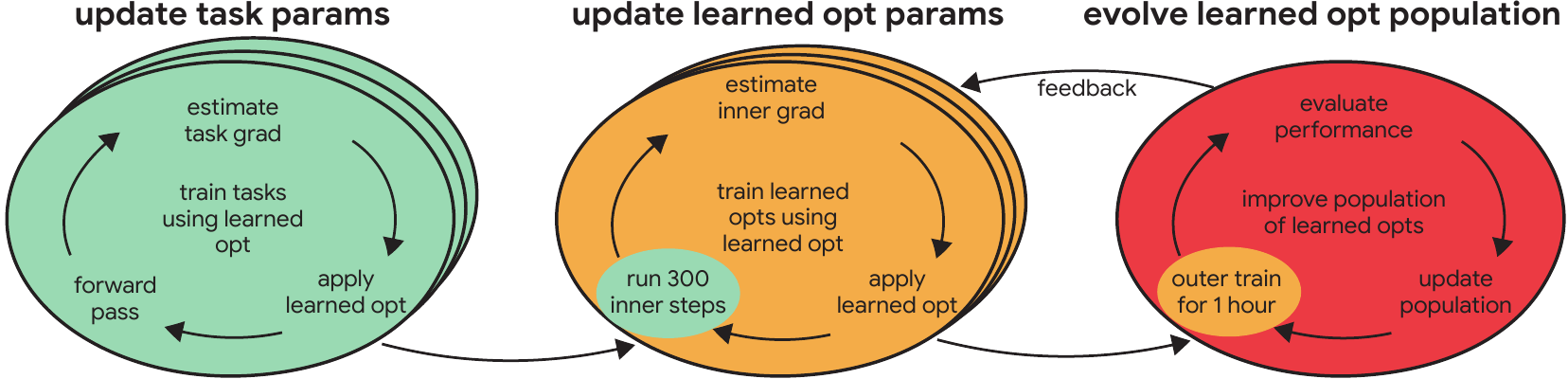}
    \vspace{-5pt}
    \caption{Schematic of the three nested layers of learning when training a learned optimizer. Each subsequent layer uses many iterations of the previous layer for each update. In the base layer (left), we train some target model using gradient descent with a learned optimizer. In the middle layer we update the learned optimizer parameters, using a learned optimizer, to perform better at training the target tasks. In the final layer (right), we employ population based training to guide the learned optimizer in the middle layer used to train the learned optimizers.
    \label{fig:schematic}
    }
\end{figure}

\section{Method and Experiment}
\begin{figure}[th]
    \centering
    \includegraphics[width=0.45\textwidth,height=15em]{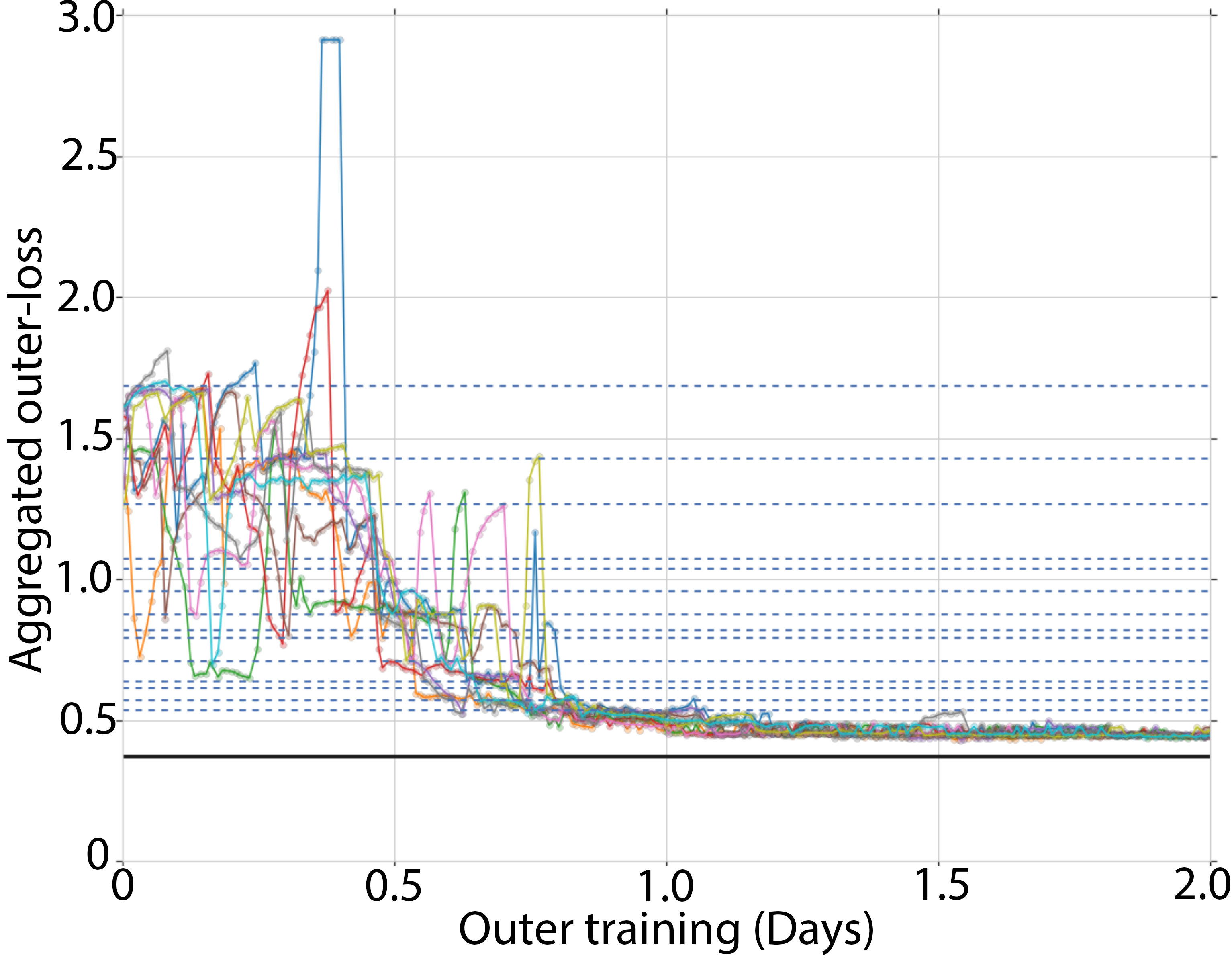}\qquad
    \includegraphics[width=0.45\textwidth,height=15em]{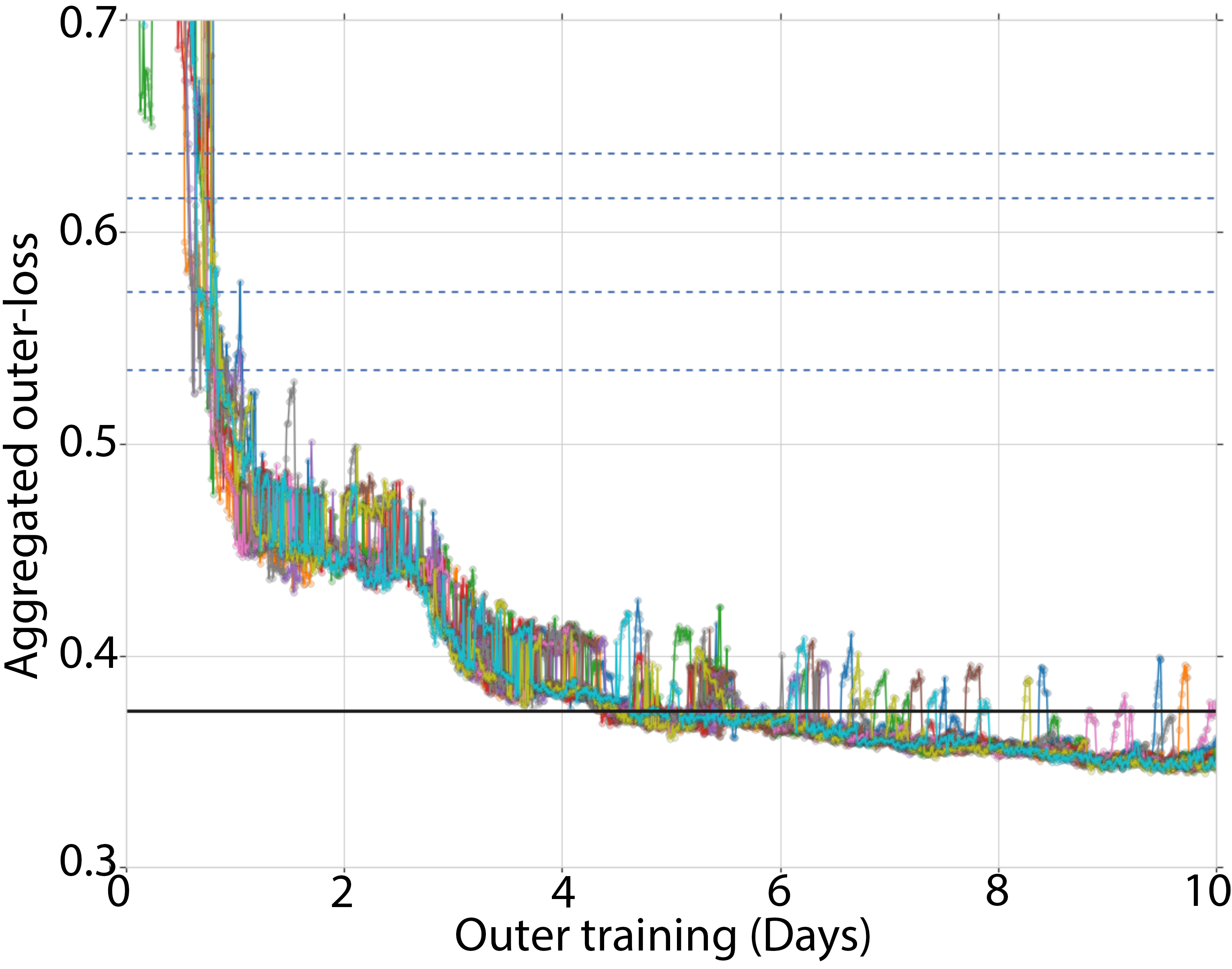}
    \vspace{-5pt}
    \caption{Training curves showing learned optimizer performance. In color we show each member of the population. Each point consists of evaluating a given learned optimizer on 100 tasks for 10,000 inner-steps then computing the mean over tasks and normalized learning curves. We show performance of fixed learning rate Adam optimizers in dashed blue lines, with each line showing a different learning rate sampled logarithmically between $10^{-6}$ and $10^1$ every half order of magnitude.
    In black we show learning rate tuned Adam with tuning done independently per task.
    \textbf{Left:} Early in outer training we find poor training as all optimizers are randomly initialized.
    \textbf{Right:} Later in training, outer-training speed increases as our learned optimizers learn to optimize, and thus optimize themselves faster.
    \label{fig:result}
    }
    \vspace{-5pt}
\end{figure}

We train neural-network parameterized learned optimizers, with an architecture based on that in \citet{metz2020tasks}. 
The outer-loss used to train these learned optimizers consists of a large diversity of different problem types (language modeling with RNNs, image classification with CNNs, and more) \citep{metz2019using}. 
These problems were designed to capture enough variation over different types of optimization problems so as to facilitate learning on the meta-problem of optimizing the learned optimizer. Put another way, being able to effectively optimize the problems in this outer loss is sufficient to enable the learned optimizer to effectively train itself.
At this point, this set of tasks is likely larger than needed to enable self-training but due to compute costs was all we explored.

Instead of training a single model with a hand designed optimizer (e.g. Adam), we initialize a population of 10 randomly initialized learned optimizers.
At initialization, we randomly select models from this population to optimize other models.
To improve this population, we employ evolution---in particular 
population based training \citep{jaderberg2017population}.
Approximately every 3 hours of outer training we compare the performance of two randomly selected optimizers evaluated on 100 tasks from the target training distribution and select the best performing learned optimizer parameters as well as the learned optimizer being used to train these parameters (the outer-optimizer).
With a 50\% chance, we randomly select a different outer-optimizer from the population of learned optimizers and with 10\% chance we re-sample length of each truncation, how many steps we unroll our optimizer before computing gradients, to be either 200, 300, 400, or 800. We run this population for 10 days on a ~500K CPU core cluster. Results are shown in Figure \ref{fig:result}.

Early in training, we find our learned optimizers either diverge, or learn slowly.
This is natural as at this point all of the optimizers used to train the learned optimizes are randomly initialized.
After a number of days our population of optimizers start to learn to optimize resulting in learned optimizers that are on par with Adam using the same fixed learning rate for all evaluation tasks.
This in turn accelerates outer-training and ultimately produces learned optimizers that outperform per problem learning rate tuned Adam on the training distribution of tasks (a strong baseline on this outer-dataset).
\section{Related Work}

Online meta-learning methods such as hyper-gradient methods, \citep{xu2018meta, xu2020meta}, or population methods \citep{jaderberg2017population, jaderberg2019human} also employ learning procedures that accelerate learning.
These systems usually operate on hyperparameters, and aim to improve performance after some number of steps of learning.
Not only do these systems limit what can be learned (e.g. modification of hyperparameters instead of full learning algorithms), they limit {\emph how} this learning occurs which limits the speed of learning.

Learned optimizers have been explored in \citep{bengio1990learning, andrychowicz2016learning, Bello17, wichrowska2017learned, li2018learning, metz2019understanding, metz2020tasks}. Using fixed learned optimizers to train new learned optimizers was also explored in \citet{metz2020tasks} but this was done in an offline, as opposed to online, fashion. Leveraging a population not only speeds up the rate of learning speedup, it also stabilizes training \citep{jaderberg2017population}.

Other work proposes or discusses self learning systems. \citet{schmidhuber1987evolutionary} explores using genetic programming 
to improve itself. \citet{schmidhuber2007godel} explores learning proof based learning algorithms that update their own prover with provably useful improvements. Finally, \citet{clune2019ai} advocates for similar learning systems by learning architecture, learning algorithms, and environments.

\section{Discussion}
In this work we demonstrate one of the first machine learning systems which accelerates its own learning by training itself.
We hope this serves as inspiration for future systems which improve themselves in a positive feedback loop.
As the tasks on which we employ machine learning become more complex, we believe solving problems by building positive feedback loops will enable a faster rate of progress than hand designed methods.

\begin{ack}
  We would like to thank
  Liam Fedus,
  Chip Huyen,
  Diogo Moitinho de Almeida,
  Timothy Nguyen,
  Ben Poole, 
  Alec Radford,
  Ruoxi Sun,
  Wojciech Zaremba,
  and the rest of the Brain team for valuable discussions on this work.

\end{ack}

\bibliography{iclr2019_conference}
\bibliographystyle{iclr2019_conference}

\end{document}